# Deep reinforcement learning methods for structure-guided processing path optimization

Johannes Dornheim[1] · Lukas Morand[2] · Samuel Zeitvogel[1] · Tarek Iraki[1] · Norbert Link[1] · Dirk Helm[2]

## Abstract

A major goal of materials design is to find material structures with desired properties and in a second step to find a processing path to reach one of these structures. In this paper, we propose and investigate a deep reinforcement learning approach for the optimization of processing paths. The goal is to find optimal processing paths in the material structure space that lead to target-structures, which have been identified beforehand to result in desired material properties. There exists a target set containing one or multiple different structures, bearing the desired properties. Our proposed methods can find an optimal path from a start structure to a single target structure, or optimize the processing paths to one of the equivalent target-structures in the set. In the latter case, the algorithm learns during processing to simultaneously identify the best reachable target structure and the optimal path to it. The proposed methods belong to the family of model-free deep reinforcement learning algorithms. They are guided by structure representations as features of the process state and by a reward signal, which is formulated based on a distance function in5 the structure space. Model-free reinforcement learning algorithms learn through trial and error while interacting with the process. Thereby, they are not restricted to information from a priori sampled processing data and are able to adapt to the specific process. The optimization itself is model-free and does not require any prior knowledge about the process itself. We instantiate and evaluate the proposed methods by optimizing paths of a generic metal forming process. We show the ability of both methods to find processing paths leading close to target structures and the ability of the extended method to identify target-structures that can be reached effectively and efficiently and to focus on these targets for sample efficient processing path optimization.

## Introduction

Manufacturing processes usually influence the internal structure of materials and the resulting material properties. Conversely, this means that tailor-made material properties can be set via the manufacturing process. The relations between processing, structure, property, and performance of materials are represented in Olson (1997) as a three-link chain model which is depicted in Fig. 1. The inversion of this three-link chain model reveals a workflow for the design of new materials and processes, driven by desired material properties and the workpiece performance. The present paper focuses on the last link of the model: The identification of optimal pro-cessing paths in a given multi-step manufacturing process aiming to produce semi-finished products or components with tailored-properties. For this purpose, a deep reinforcement learning approach is proposed for optimal path-finding, which is guided by a reward function defined in the native space of the material structures. The method is generic for any set of target-structures and therefore independent of specific target-properties and property types, as long as the material structure is correlated to the target-properties.

A material property usually has a one-to-many relation to structures, which means that a set of structures is equivalent with respect to the desired material properties. The aim of processing path optimization is then to find a path from an initial structure to one of the equivalent target-structures. In the present contribution, we propose an algorithm to find optimal paths to a single target-structure and an extended algorithm to solve problems with multi-equivalent target-structures. The contribution of *structure-guided processing path optimization* to the inverse optimization of processing-structure-property relationships is depicted in Fig. 2. The

✉ Johannes Dornheim
johannes.dornheim@mailbox.org

1 Institute Intelligent Systems Research Group, Karlsruhe University of Applied Sciences, Karlsruhe, Germany

2 Fraunhofer Institute for Mechanics of Materials IWM, Freiburg, Germany

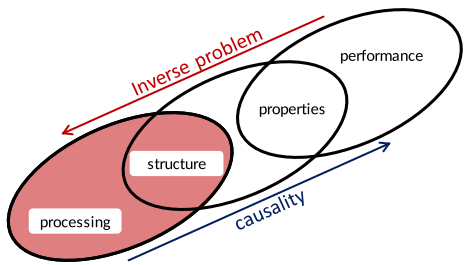

**Fig. 1** Three-link chain model following (Olson 1997). The focus of this publication is the highlighted processing-structure link. The proposed approach complements methods for inverse optimization of structure-property relationships

theory of the methods proposed in this publication apply to structures of all scales. In the materials design context, the microstructure is of special interest, which is why we focus on the processing of these in our use-cases. In the following, we use the term microstructure without loss of generality.

The proposed method is instantiated and evaluated with a synthetic metal forming sample process. The process consists of a sequence of forming steps, where a uniaxial displacement is applied to the material with a certain direction and magnitude being chosen by the algorithm at each step. A special form of this generic process is e.g. the *open die forging* process. We simulate the evolution of crystallographic texture as one of the major microstructural features that is affecting macroscopic material properties. In process-integrated online reinforcement learning, the reinforcement agent directly interacts with the process: The optimization algorithm obtains the current crystallographic texture and determines the next deformation step during the process execution. The proposed algorithm can be set in series to any method for inverse structure-property mappings. Recent examples of such methods in metal forming are described for example in Liu et al. (2015); Paul et al. (2019). The core of both approaches is a supervised learning-based identification scheme for regions in microstructure space yielding a set of optimal or near-optimal microstructures with respect to given desired properties.

## Related work

A general approach for materials design and optimal processing is described by the *Microstructure Sensitive Design* (MSD) framework, which goes back to the publication of Adams et al. Adams et al. (2001). An extensive review of publications in the context of the MSD framework and its applications can be found in Fullwood et al. (2010). Central to MSD is the description of microstructures using *n*-point correlation functions transformed into a spectral representation. After this transformation, microstructures are defined by single points in a high-dimensional spectral space and the set of microstructures obtaining specific target-properties can be identified in the form of points lying on iso-property hyperplanes. The overall framework of MSD is described in Fullwood et al. (2010) as a seven-step process starting from the definition of the principal properties (step one) and ending with the identification of processing paths to reach a target-microstructure from a given starting microstructure (step seven). While a huge corpus of publications exists for the first steps, only a few publications can be found focusing on processing path optimization in the context of materials design (Shaffer et al. 2010; Li et al. 2007; Acar and Sundararaghavan 2016, 2018; Sundar and Sundararaghavan 2020; Tran et al. 2020). Following the MSD approach, in Shaffer et al. (2010), a so-called *texture evolution network* is built based on a priori sampled processing paths. The *texture evolution network* is a directed tree graph with crys-tallographic textures as vertices and process steps as edges. Once the graph is built, it is mapped to property space by a microstructure-property mapping. In property space, graph search algorithms are used to find processing paths from a given graph-node (which is denoting the starting microstructure) to a target-node (which is denoting a microstructure with desired properties). The sequence of edges passed is then the aimed processing path. In Li et al. (2007), crystallographic texture evolution models for rolling and uniaxial tension processes are used to create a network in the space of microstructures, consisting of processing-streamlines. Linear differential transition functions are derived from the continuity equation (following Bunge and Esling 1984) and are used to describe the microstructure evolution for each process. Similar to the *texture evolution networks* approach, the a priori calculated streamlines can be used as a process evolution model and search algorithms can be used to query the model. In Acar and Sundararaghavan (2016), one-step deformation processes are optimized to reach an element of a set of target-microstructures. Tension, compression, and various shear processes are represented by so-called *process planes*. These are per process principal component analysis (PCA) projections of microstructures that are reachable by the specific process. The optimization is performed by searching for the process plane which best represents one of the target-microstructures. In Acar and Sundararaghavan (2018) this approach is extended by calculating PCA projections also for two- and three-step processes. In Sundar and Sundararaghavan (2020) variational autoencoders are used to represent processed microstructures in a lower-dimensional space. A priori sampled microstructures with associated processing paths are stored in a database. For a given target-microstructure, microstructures with known processing paths are then identified in the database by using a distance-function defined in the low-dimensional space. The method is applied to a data set, sampled from parametrized tension, compression and rolling processes with up to four process steps. The approach in Tran et al. (2020) combines Bayesian optimization and kinetic Monte Carlo simulations to optimize welding process parameters and a grain growth process with the objective defined in the microstructure space.

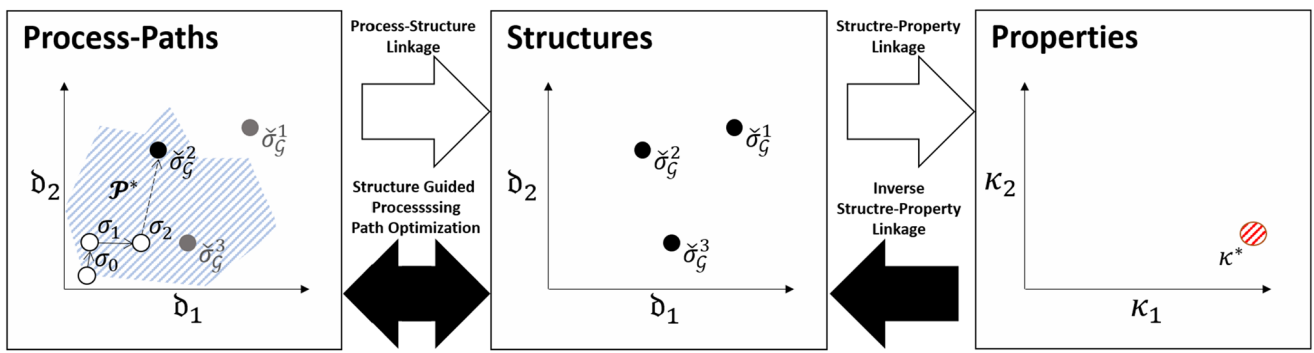

**Fig. 2** Overview of *structure-guided processing path optimization* in the context of inverse structure-property mappings. The direct three-link chain (Processes-Structures-Properties) is depicted by white arrows; its inversion by black arrows. Desired properties $\mathcal{K}^*$ are mapped to a set of target-structures $\check{\sigma}^g_{\mathcal{G}} \in \mathcal{G}$. Structures $\sigma$ are described by features $\mathfrak{d}(\sigma)$. For a given initial structure $\sigma_0$, the aim of *structure-guided processing* *path optimization* is then to find an efficient sequence of processing-steps to one of the target-structures from $\mathcal{G}$. As indicated by the shaded process-area in the structure feature-space, not all structures from $\mathcal{G}$ are necessarily reachable by the process. The optimization is guided by the structure features and a reward function defined in the space of structures

The majority of the methods discussed above are based on a search-algorithm, either directly searching in a set of a priori sampled processing paths (Shaffer et al. 2010; Acar and Sundararaghavan 2016, 2018; Sundar and Sundararaghavan 2020) or in a generalized function derived from a priori processing paths (Li et al. 2007). Thereby, they implicitly rely on the assumption, that the sampled set is representative and fully describes the microstructure space. In contrast, only the approach presented in Tran et al. (2020) actively samples a process simulation. When it comes to the process-path length, most of the algorithms optimize single-step processes (Acar and Sundararaghavan 2016; Tran et al. 2020) or processes with only few processing-steps (up to six processing-steps) (Sundar and Sundararaghavan 2020; Acar and Sundararaghavan 2018). Except for Acar and Sundararaghavan 2016 and Acar and Sundararaghavan (2018), the discussed methods do not explicitly consider the existence of sets of equivalent microstructure solutions for a specified target-property (Table 1).

**Table 1** Table of the most important symbols used throughout the paper

| Symbol | Quantity |
|---|---|
| | *Reinforcement learning* |
| $S$ | State-Space, Set of states $s$ |
| $A$ | Action-Space, Set of actions $a$ |
| $t \in \mathbb{N}$ | Time-step within an episode |
| $K \in \mathbb{R}^+$ | Maximum length of episode/Processing path |
| $P(s_{t+1}\|s_t, a_t)$ | Probability of the transition to $s_{t+1}$ when taking $a_t$ in $s_t$ |
| $R(s_t, a_t, s_{t+1})$ | Reward for the transition from $s_t$ to $s_{t+1}$ via $a_t$ |
| $\gamma \in [0, 1]$ | Discount factor |
| $\pi, \pi^*$ | (possibly stochastic) control policy, optimal $\pi$ |
| $\mathcal{V}_\pi : S \to \mathbb{R}$ | Expected future reward for $\pi$ |
| $\mathcal{V}^* : S \to \mathbb{R}$ | Expected future reward for $\pi^*$ |
| $\mathcal{Q}^* : S \times A \to \mathbb{R}$ | Expected future reward for $\pi^*$ |
| $x$ | Experience tuple $(s_t, a_t, s_{t+1}, R)$ |
| $\mathcal{D}$ | Replay memory, consisting of experience tuples $x$ |
| $\alpha \in [0, 1]$ | Q-Learning learning rate |
| $\boldsymbol{\theta}, \boldsymbol{\theta}^-$ | Function approximation weights |
| | *Application* |
| $\mathcal{S}$ | Space of microstructures $\sigma, \sigma_t, \sigma^*$ |
| $\mathcal{K}$ | Space of material properties $\kappa$ |
| $\mathcal{P}$ | Process path |
| $\mathcal{P}^*$ | Optimal process path |
| $\mathcal{G}$ | Set of equivalent target-microstructures $\tilde{\sigma}_\mathcal{G}$ |
| $\mathcal{S}_\mathcal{P} \subseteq \mathcal{S}$ | Reachable Microstructures |
| $d_\sigma : \mathcal{S} \times \mathcal{S} \to \mathbb{R}$ | microstructure distance-function |
| $f(h)$ | Orientation Density Function |
| | *Taylor model* |
| $T^{(i)}$ | Cauchy stress of $i$th crystal |
| $\boldsymbol{F}$ | Deformation gradient |
| $\boldsymbol{L}$ | Velocity gradient |
| $\boldsymbol{R}$ | Rotation matrix |
| $E$ | Young's modulus |

## Contribution

We propose a model-free deep reinforcement learning algorithm for *structure-guided processing path optimization* and an extended algorithm for sample-efficient processing path optimization in the case of various equivalent target-microstructures. The proposed algorithms can optimize processing paths of arbitrary length (in our application study we evaluate on paths with up to 100 processing-steps). The proposed algorithms are not tied to a specific process model and do not need a priori samples. Instead, they query the process simulation during optimization (online). Online algorithms are not restricted to the information which is represented in a priori sampled datasets and particularly sample from regions that are frequently visited during optimization. This usually has a positive impact on sample-efficiency and the quality of optimization results (Sutton and Barto 2018).

In the context of manufacturing, reinforcement learning methods have been proposed for model-free adaptive optimization on the device level and the operational level of various manufacturing processes. Recently published work include the optimization of process control in sheet metal milling (Veeramani et al. 2019), polymerization reaction systems (Ma et al. 2019), laser welding (Günther et al. 2016) and in deep drawing (Dornheim et al. 2019). Operational optimization objects are amongst others material flow in industrial mining (Kumar et al. 2020), preventive maintenance scheduling of flow line systems (Wang et al. 2016) and job shop scheduling (Kuhnle et al. 2020).

The algorithms proposed in this paper are based on *deep Q networks* (DQN) (Mnih et al. 2015) with several enhancements (Schaul et al. 2015b; Van Hasselt et al. 2016; Wang et al. 2015). We extend DQN by a goal-prioritization and data-augmentation mechanism to exploit the fact that several target-microstructures are equivalent regarding the objective. The basis for these extensions is the generalization of the reinforcement learning value functions to include the target-microstructures, as it is done in the *universal value function approximator* approach (Schaul et al. 2015a). The data-augmentation mechanism we propose is inspired by *hindsight experience replay* (Andrychowicz et al. 2017). We utilize *potential based reward shaping* (Ng et al. 1999) for sample-efficient learning.

The contribution of this paper is four-fold:

(1) The proposition of a deep reinforcement-learning algorithm to solve *single-target structure-guided processing path optimization* problems.
(2) The extension of the proposed single-target algorithm to a novel deep reinforcement-learning algorithm to efficiently solve multi-equivalent-goal optimization problems.
(3) The proposition of a histogram-based distance-function for orientation densities to describe the crystallographic texture.
(4) An extensive quantitative evaluation of our algorithms on a generic metal forming process.

To evaluate the proposed methods, we use an efficient mean-field crystal plasticity model of Taylor type for the study presented in this paper. The target of the optimized process is the crystallographic texture in a single-phase polycrystalline material. Nevertheless, due to the generic nature of our method, it can also be combined with more accurate but computationally expensive full-field simulation approaches, see for example (Roters et al. 2010; Eisenlohr et al. 2013). Furthermore, it is not restricted to optimize crystallographic texture but can be applied to any kind of microstructural or material quantity.

To guarantee the reproducibility of our results and to encourage the development of *structure-guided processing path optimization* methods, we publish the used simulation wrapped by an openAI Gym (Brockman et al. 2016) reinforcement learning environment[1].

## Paper structure

The paper is structured as follows: In Sect. 2 we introduce the fundamental basics of the reinforcement learning approach and the material model used for our evaluation case-study. Then, in Sect. 3 our proposed algorithms for single-target and multi-equivalent-target *structure-guided processing path optimization* are described. Section 4 describes the application case study that is used to evaluate our methods. In Sect. 5, evaluation results are reported. Finally, the research results are summarized in Sect. 6.

## Background

### Markov decision process and dynamic programming

Markov decision processes (MDP) are defined by the 5-tuple $(S, A, P, R, \gamma)$ and build the formal basis of the problem definition for the algorithms of reinforcement learning and of dynamic programming. $S$ is the set of all states $s$ of the system, and $A$ is the set of allowed control actions $a$. $P(s_{t+1}|s_t, a_t)$ is the probability of transition to the successor state $s_{t+1}$ when applying action $a_t$ in state $s_t$. Due to the Markov property, the one-step transition probability function $P$ is sufficient to completely describe the transition dynamics of the decision process. $R(s_t, a_t, s_{t+1})$ is the reward function,

---

delivering a real-valued reward signal, when a transition from $s_t$ to $s_{t+1}$ takes place, triggered by action $a_t$ at time step $t$. The discount factor $\gamma \in [0, 1]$ is the degree of future reward discounting.

For a given control policy $\pi$, a possibly stochastic mapping from states to actions $A \sim \pi(S)$ and time-index $t$, the state value function

$$\mathcal{V}_\pi(s_t) = \mathbb{E}_{P,\pi}\Big[R(s_t, a_t, s_{t+1}) + \gamma \mathcal{V}_\pi(s_{t+1})\Big], \tag{1}$$

denotes the expected discounted future reward when applying the control policy $\pi$ from state $s_t$ onwards, where $a_t \sim \pi(s_t)$ and $s_{t+1}$ is drawn according to $P$. The MDP is considered to be solved when an optimal control policy $\pi^*$ is found, for which the expected future reward is maximized:

$$\pi^*(s) = \arg\max_{a \in A} \mathbb{E}_P\Big[\mathcal{V}^*(s_{t+1})\Big], \tag{2}$$

for all $s \in S$, where $\mathcal{V}^* = \mathcal{V}_{\pi^*}$. As it can be seen in (2), $\pi^*$ can be extracted from $\mathcal{V}^*$. Consequently, finding $\pi^*$ is equivalent to finding $\mathcal{V}^*$. When the transition function $P$ is known in advance, $\mathcal{V}^*$ can be determined by dynamic programming methods.

## Deep reinforcement learning

If the transition function $P$ is not known, methods from reinforcement learning can be used to find the optimal policy in an iterative *learning-by-doing* fashion, by interacting with the processes and, based on the experiences made, deriving the optimal policy $\pi^*$. In function-based reinforcement learning, expected future reward functions, like the value function (1), are modeled and the model is used to solve the MDP. The objective of Q-learning based methods is to identify the optimal Q-function $\mathcal{Q}^*$, also named *state-action value function*:

$$\mathcal{Q}^*(s_t, a_t) = \mathbb{E}_P\Big[R(s_t, a_t, s_{t+1}) + \gamma \mathcal{Q}^*(s_{t+1}, \pi^*(s_{t+1}))\Big]. \tag{3}$$

Unlike the value function from (1), the Q-function captures the one-step system dynamics by taking the actions into account. In Q-Learning, the approximation of the optimal Q-function is updated based on experience-tuples $x_t = (s_t, a_t, s_{t+1}, R(s_t, a_t, s_{t+1}))$ gathered in each time-step by interacting with the process. The update is as follows:

$$\mathcal{Q}'(s_t, a_t) \leftarrow \mathcal{Q}(s_t, a_t) + \alpha(\delta_{\mathcal{Q}}(x_t)), \tag{4}$$

where $\alpha \in [0, 1]$ is the Q-learning rate and $\delta_{\mathcal{Q}}(x_t)$ is the *time difference error*, defined by

$$\begin{aligned} \delta_{\mathcal{Q}}(x_t) = {} & R(s_t, a_t, s_{t+1}) \\ & + \gamma \max_{a_{t+1} \in A} \mathcal{Q}(s_{t+1}, a_{t+1}) - \mathcal{Q}(s_t, a_t). \end{aligned} \tag{5}$$

The Q-function is guaranteed to converge to $\mathcal{Q}^*$ if the explorative policy, which is used during learning guarantees that the probability of taking an action $a$ in state $s$ is non-zero for every $(s, a) \in S \times A$ (Watkins and Dayan 1992). Often an $\epsilon$-greedy policy is used for this purpose, which acts randomly in an $\epsilon$ fraction of time-steps and greedy (according to the current Q-function estimation) in all other cases (Sutton and Barto 2018).

Pure Q-learning (without function approximation) is subject to the *curse of dimensionality*, like other algorithms from reinforcement learning and dynamic programming. This is leading to difficulties in terms of the data complexity required for sampling high dimensional state-action spaces. To overcome these difficulties, in real-world applications, Q-learning is often combined with supervised learning of an approximation of the Q-function. When artificial neural networks are used for function approximation, the Q-function update learning rate $\alpha$ is dropped in favor of the back-propagation optimizer learning rate. The training target for a specific experience-tuple $x_t$ is then

$$Y_{\mathcal{Q}} = R(s_t, a_t, s_{t+1}) + \gamma \max_{a_{t+1} \in A} \mathcal{Q}(s_{t+1}, a_{t+1}, \boldsymbol{\theta}), \tag{6}$$

where $\mathcal{Q}(s, a, \boldsymbol{\theta})$ is the Q-function approximation, represented by an artificial neural network with weights $\boldsymbol{\theta}$. It is important to note that the theoretical convergence guarantees from pure Q-learning cease to hold when function approximators are used.

In this paper we use the *deep Q Networks* (DQN) algorithm (Mnih et al. 2015) as the core reinforcement learning algorithm. DQN combines Q-function approximation using a deep artificial neural network with *experience replay* (Lin 1992). In typical reinforcement learning tasks, successive states in the process trajectory are highly correlated. Due to this fact, updating the artificial neural network weights in the order of incoming experiences $x_t$ breaks the i.i.d. assumption and leads to a phenomenon called *catastrophic forgetting*. This means that early experiences that may become useful later on in the training process are buried under later experiences. Experience replay addresses this, by storing the experiences in a so-called *replay memory* $\mathcal{D} = [e_0, e_1, \dots, e_n]$ which is used to sample data for mini-batch updates of the DQN in a stochastic fashion during learning. Due to the dependency of training targets on the approximated Q-function with constantly updated parameters $\boldsymbol{\theta}$ as shown in (6), learning can still be unstable. In DQN, *target-parameters* $\theta^-$, used for target-calculations, are therefore decoupled from *online-parameters* $\theta$, which are continuously updated and used to derive the explorative policy $\pi$. The target-parameters are kept fixed for $n_\theta$ steps and are then updated by copying the online-parameters $\theta^- \leftarrow \theta$. The per-sample training target is then

$$Y_{DQN} = R(s_t, a_t, s_{t+1}) + \gamma \max_{a_{t+1} \in A} \mathcal{Q}(s_{t+1}, a_{t+1}, \theta^-). \tag{7}$$

Further DQN enhancements have been shown to improve learning performance and stability. The enhancements we employ are *prioritized experience replay* (PER) (Schaul et al. 2015b), *double Q-learning* (Van Hasselt et al. 2016) and *dueling Q-learning* (Wang et al. 2015). In standard DQN, experiences are sampled uniformly from the replay memory $\mathcal{D}$ to form mini-batches for network training. As the name PER suggests, some experiences are prioritized during sampling. PER is based on the assumption that experiences with a higher time-difference error $\delta_{\mathcal{Q}}$ hold more information regarding the optimization task. To ensure that the prioritization does not lead to a loss of diversity in the training data, PER uses a stochastic prioritization. The PER hyperparameter $\alpha_{PER} \in \mathbb{R}_0^+$ determines the influence of the prioritization, where $\alpha_{PER} = 0$ corresponds to uniform sampling. Any deviation from uniform sampling is in conflict with the i.i.d. assumption of supervised learning. In reinforcement learning this becomes noticeable especially towards the end of the learning process. The resulting bias is corrected in PER by using *weighted importance-sampling*. The amount of initial importance sampling correction $\beta$ is specified by a second PER hyperparameter $\beta_0$ and then linearly approaches 1.0 during training. For an in-depth explanation of the stochastic prioritization and importance sampling, we refer to Schaul et al. (2015b).

When using Q-function approximation, the maximization over estimated action values in the calculation of $\delta_{\mathcal{Q}}$ (5) can lead to a systematic overestimation of expectation values (Thrun and Schwartz 1993). *Double Q-learning* (Van Hasselt et al. 2016) reduces such an overestimation by decoupling the maximation of the action value from the action evaluation. The maximization is then done by the online-network, while the evaluation is still conducted by the network with the target parameters. The resulting per-sample target is

$$\begin{aligned} Y_{DDQN} = R + \gamma\, \mathcal{Q}(s_{t+1}, \\ \arg\max_{a_{t+1} \in A} \mathcal{Q}(s_{t+1}, a_{t+1}, \theta), \theta^-), \end{aligned} \tag{8}$$

where $R = R(s_t, a_t, s_{t+1})$.

In *Dueling Q-learning* a decomposition of the Q-network architecture into two separate streams is proposed. The decomposition is based on the relationship $\mathcal{Q}_\pi(s, a) = $ $\mathcal{V}_\pi(s), \mathcal{A}_\pi(s, a)$, where $\mathcal{V}_\pi$ is the *state-value function* from (1) and $\mathcal{A}_\pi$ is the so-called advantage-function, the delta of the value of a given state $s$ and the value of applying action $a$ in $s$. Dueling Q-learning is compatible with PER and *double Q-Learning*. For details we refer to Wang et al. (2015).

## Taylor-type material model

For the purpose of this work, a Taylor-type crystal plasticity model is used as is described in Kalidindi et al. (1992). The Taylor-type model is based on the assumption that all crystals in a polycrystalline aggregate experience the same deformation and does not consider spatial relations and morphological texture. In addition to the classical Taylor model, the stress response of the material is given by the average volume-weighted stress over all $n$ crystals

$$\overline{\boldsymbol{T}} = \frac{1}{V} \sum_{i=1}^{n} \boldsymbol{T}^{(i)} V^{(i)}, \tag{9}$$

where $\boldsymbol{T}$ denotes Cauchy stress tensor. Based on the multiplicative decomposition of the deformation gradient in its elastic and plastic parts $\boldsymbol{F} = \boldsymbol{F}_e \cdot \boldsymbol{F}_p$, the stress tensor in the intermediate configuration can be derived for a single crystal:

$$\boldsymbol{T}^* = \frac{1}{2} \, \mathbb{C} : (\boldsymbol{F}_e^T \cdot \boldsymbol{F}_e - \boldsymbol{I}), \tag{10}$$

where $\boldsymbol{I}$ is the second order identity tensor, $\mathbb{C}$ is the fourth order elastic stiffness tensor, and the $\cdot$ and $:$ operators stands for the dot and the double dot product. For cubic symmetry, $\mathbb{C}$ has only three independent parameters: $C_{11}$, $C_{12}$, and $C_{44}$. For iron single crystals, these can be set to 226.0, 140.0, and 116.0 GPa, respectively (Frederikse 2008). The relation between $\boldsymbol{T}$ and $\boldsymbol{T}^*$ is given by

$$\boldsymbol{T}^* = \boldsymbol{F}_e^{-1} \cdot (\det(\boldsymbol{F}_e)\, \boldsymbol{T}) \cdot \boldsymbol{F}_e^{-\top}. \tag{11}$$

The evolution of the plastic deformation is considered on the basis of the plastic part of the velocity gradient $\boldsymbol{L}_p$:

$$\boldsymbol{L}_p = \dot{\boldsymbol{F}}_p \cdot \boldsymbol{F}_p^{-1}. \tag{12}$$

The flow rule is defined as the sum of the shear rates $\dot{\gamma}$ on every active slip system $\eta$ (Rice 1971):

$$\boldsymbol{L}_p = \sum_\eta \dot{\gamma}^{(\eta)} \boldsymbol{m}^{(\eta)} \otimes \boldsymbol{n}^{(\eta)}. \tag{13}$$

The slip systems are defined via the slip plane normal $\boldsymbol{n}$ and the slip direction $\boldsymbol{m}$. For body-centered cubic crystals, the slip systems are given by the Miller indices $\{110\}<111>$ and $\{112\}<111>$. For simplicity and with adequate accuracy, the slip system family {123}<111> is not incorporated. To solve (13), a phenomenological approach is used. The shear rates are given by the power-law as it is described for example in Asaro and Needleman (1985):

$$\dot{\gamma}^{(\eta)} = \dot{\gamma}_0 \left| \frac{\tau^{(\eta)}}{r^{(\eta)}} \right|^{1/m} \text{sign}(\tau^{(\eta)}), \tag{14}$$

where $r^{(\eta)}$ is the slip system resistance, and $\dot{\gamma}_0$ and $m$ stand for the reference shear rate and the rate sensitivity and are set to $0.001 \text{ sec}^{-1}$ and $0.02$, respectively (Zhang et al. 2016). The resolved shear stress $\tau^{(\eta)}$ is defined by Schmid's law

$$\tau^{(\eta)} = ((\boldsymbol{F}_e^T \cdot \boldsymbol{F}_e) \cdot \boldsymbol{T}^*) : (\boldsymbol{m}^{(\eta)} \otimes \boldsymbol{n}^{(\eta)}). \tag{15}$$

To model the hardening behavior, a description is used which can be found for example in Baiker et al. (2014) and Pagenkopf et al. (2016). The hardening model is of an extended Voce-type (Tome et al. 1984):

$$\hat{\tau}^{(\eta)} = \tau_0 + (\tau_1 + \vartheta_1 \Gamma)(1 - e^{-\Gamma \vartheta_0 / \tau_1}), \tag{16}$$

in which $\tau_0, \tau_1, \vartheta_0$, and $\vartheta_1$ are material dependent parameters that are calibrated to experimental data of DC04 steel[2] and therefore set to $\tau_0 = 90.0$ MPa, $\tau_1 = 32.0$ MPa, $\vartheta_0 = 250.0$ MPa, and $\vartheta_1 = 60.0$ MPa. The accumulated plastic shear $\Gamma$ is given by

$$\Gamma = \int_0^t \sum_\eta \left| \dot{\gamma}^{(\eta)} \right| dt. \tag{17}$$

The evolution of the slip system resistance can be calculated via

$$\dot{r}^{(\eta)} = \frac{d\hat{\tau}^{(\eta)}}{d\Gamma} \sum_\zeta \hat{q}_{\eta\zeta} |\dot{\gamma}^{(\zeta)}|, \tag{18}$$

where $\hat{q}_{\eta\zeta}$ is a matrix that describes the ratio between latent and self-hardening. Having two slip system families, then $\hat{q}_{\eta\zeta}$ is composed out of the parameters $\hat{q}_1$ and $\hat{q}_2$, while the diagonal elements are equal to 1.0. $\hat{q}_1$ represents the latent hardening of coplanar slip systems and $\hat{q}_2$ of non-coplanar slip systems. Both values are set to 1.4 (Asaro and Needleman 1985).

### Representation of crystallographic textures

In this work, the crystallographic texture is described by a one-point correlation function, namely the orientation distribution function (ODF). The ODF is commonly defined as a

continuous and non-negative function $f(h)$ with

$$f(h)dh = \frac{V(h)}{V} \text{ and } \int_{SO(3)} f(h)dh = 1, \tag{19}$$

where $h$ is a point in $SO(3)$ and $V(h)$ is the volume of $h$ in the orientation space. To describe the ODF of metallic materials, conditions for crystal and sample symmetry can be incorporated. Due to these symmetry conditions, various regions of $SO(3)$ are equivalent. Therefore, orientations can be mapped to a *fundamental zone*, such that the ODF described by the mapped orientations $f(\tilde{h})$ is physically indistinguishable from $f(h)$ (Bunge 2013).

For function approximation in the context of reinforcement learning a low dimensional state-representation that describes the state adequately by ordered features is preferred. These requirements can be fulfilled by approximating the ODF using *generalized spherical harmonics* (GSH) (Bunge 2013) and taking its coefficients as state-representation. For this reason, GSH coefficients are popular ODF descriptors in machine learning and data mining applications (e.g. Fullwood et al. 2010; Shaffer et al. 2010; Li et al. 2007; Acar and Sundararaghavan 2018). By embedding crystal and sample symmetry, the GSH can be symmetrized to reduce the number of necessary coefficients:

$$f(h) = \sum_{l=0}^{\infty} \sum_{u=1}^{M(l)} \sum_{v=1}^{N(l)} C_l^{uv} \ddot{T}_l^{uv}(h), \tag{20}$$

where $\ddot{T}_l^{uv}$ are the symmetrized generalized harmonics functions and $C_l^{uv}$ are the corresponding coefficients. $M(l)$ and $N(l)$ describe the number of linearly independent harmonics for the crystal and sample symmetry and are defined in Bunge (2013). Eq. (20) depicts an infinite series, which is usually truncated to a certain degree based on the application of interest. We denote the GSH coefficient vector of a texture $\sigma$ with the crystal system $\Omega$ of grade $l$ by $\zeta_l^\Omega(\sigma)$. The representation of the ODF via a vector of GSH coefficients does not allow the unambiguous calculation of distances between crystallographic textures, which is a prerequisite to reach given target-microstructures. An alternative representation will therefore be proposed in Sect. 4.

## Method

### Objective

For a given target-microstructure $\breve{\sigma} \in \mathcal{S}$, *structure-guided processing path optimization* aims to find a sequence of processing-steps $\mathcal{P}^* = (a_0, a_1, ..., a_{N-1}, a_N)$, where $N <$

---

[2] experiments performed at IUL Dortmund during DFG project Graduate School 1483 (Pagenkopf 2019)

$K$ and $K$ is the maximum length of $\mathcal{P}$, that leads from an initial microstructure $\sigma_0$ to the nearest possible microstructure

$$\sigma^* = \arg\min_{\sigma \in \mathcal{S}_\mathcal{P}} d_\sigma(\sigma, \check{\sigma}), \qquad (21)$$

from the set of microstructures $\mathcal{S}_\mathcal{P} \subseteq \mathcal{S}$ which are reachable by the process, where $d_\sigma : \mathcal{S} \times \mathcal{S} \to \mathbb{R}_0^+$ is a distance function in the microstructure space $\mathcal{S}$. The optimal processing path $\mathcal{P}^*$ is the shortest among all paths from $\sigma_0$ to $\sigma^*$. Neither $\mathcal{S}_\mathcal{P}$ nor $\sigma^*$ are known in advance. Instead, an interactive microstructure changing process is given.

Typically multiple equivalent microstructures exist, which bear the desired material properties exactly or at least in a sufficiently small range. Approaches for the inversion of structure-property mappings therefore often return a set $\mathcal{G}$ of target-microstructures $\check{\sigma}_\mathcal{G} \in \mathcal{G}$ instead of a single target-microstructure $\check{\sigma}$ (Liu et al. 2015; Paul et al. 2019). The additional aim of processing path optimization is then to identify the best reachable target-microstructure $\check{\sigma}_\mathcal{G}^* \in \mathcal{G}$ and Eq. (21) extends to

$$(\sigma^*, \check{\sigma}_\mathcal{G}^*) = \arg\min_{(\sigma, \check{\sigma}_\mathcal{G}) \in \mathcal{S}_\mathcal{P} \times \mathcal{G}} d_\sigma(\sigma, \check{\sigma}_\mathcal{G}). \qquad (22)$$

Generally, processing path optimization algorithms can take two basic strategies to solve this extended task:

(a) Independently optimize paths for each $\check{\sigma}_\mathcal{G} \in \mathcal{G}$, to identify $\check{\sigma}_\mathcal{G}^*$ retrospectively.
(b) Integrate the identification of $\check{\sigma}_\mathcal{G}^*$ into the path optimization procedure.

Strategy (a) results in a straight forward procedure using a standard single-goal process path optimization algorithm for every microstructure in $\mathcal{G}$, while strategy (b) often requires adoptions of the standard single-target path optimization algorithm. If the optimization algorithm works online like the proposed approach and not on a priori process samples, the overall process efficiency is highly dependent on the number of interactions. In the case of real processes, this determines the processing time, energy, and tool wear. In the case of simulated processes, it determines the computational cost of the process simulation. Strategy (a) uses the maximum amount of resources by optimizing processing paths for each $\check{\sigma}_\mathcal{G} \in \mathcal{G}$ independently. By integrating the identification of the optimally reachable microstructure into the optimization procedure, the algorithm focuses its resources early on target-microstructures that are considered to be reachable and ignores target-microstructures that are very likely unreachable. Therefore, we propose a *multi-equivalent-goal* reinforcement learning approach, which follows strategy (b).

In the remainder of this section, we firstly propose a deep reinforcement learning approach to solve *single-goal structure-guided processing path optimization problems*,

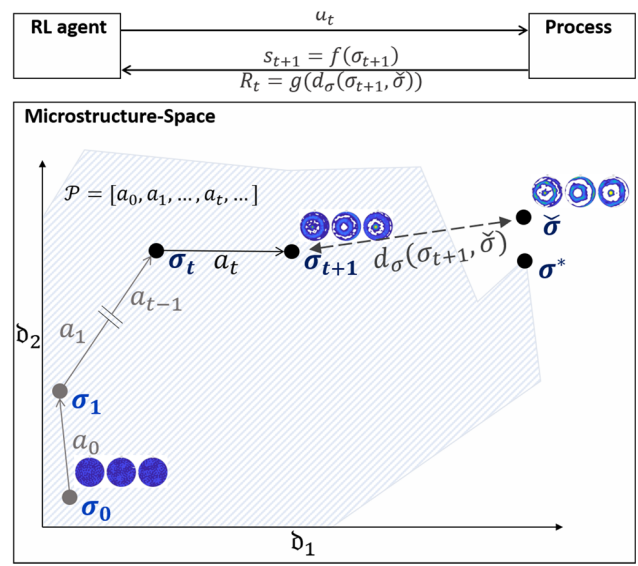

**Fig. 3** Schematic overview of *single-goal structure-guided processing path optimization*. For the target-microstructure $\check{\sigma}$, the task is to find a path of process actions $\mathcal{P} = [a_0, a_1, ....]$ that determines a microstructure $\sigma^*$, close to $\check{\sigma}$. The process is a black-box for the reinforcement learning agent. After executing $a_t$ the process emits a state description $\mathbf{s_{t+1}}$ and a reward signal $R_{t+1} \in \mathbb{R}$. The reinforcement learning agent is structure-guided in the sense that $\mathbf{s_t}$ is a description of the microstructure $\sigma_t$ and the reward-signal $R_t$ is a function $g$ of the microstructure distance $d_\sigma(\sigma_{t+1}, \check{\sigma})$. The region of reachable microstructures is depicted by the shaded area. Here, microstructures are represented by the crystallographic texture, depicted by 100, 110, and 111 pole figure plots

as depicted in Fig. 3, and then extend it to our *multi-equivalent-goal structure-guided processing path optimization* approach.

## Single-goal structure-guided processing path optimization

A microstructure changing process can be described by the set of allowed processing-steps $a \in A$, a set of process state descriptions $s \in S$ and a transition function $P(s_{t+1}|s_t, a_t)$. The state description $s_t$ at time step $t$ consists of a representation $\mathfrak{d}(\sigma_t)$ of the associated microstructure $\sigma_t$ and in some cases of additional process-state related quantities, like process-conditions or external process parameters. For the process studied in this paper, the microstructure is represented by truncated *generalized spherical harmonics*, thus $\mathfrak{d}(\sigma_t) = \zeta_l^\Omega(\sigma_t)$. The transition function $P$ is not defined in an explicit form but can be sampled by interacting with the microstructure changing process or the process simulation.

To complete the formulation of the *structure-guided processing path optimization* as a Markov decision process $(S, A, P, R, \gamma)$, a reward function $R$ and discount factor $\gamma$ is needed. The proposed reward-formulation is based on a distance function $d_\sigma$, defined in the space of microstructure descriptors. Based on $d_\sigma$, a reward function which is in accor-dance with the formulation of the optimization objective $\mathcal{P}^*$ is given as

$$R(s_t, a_t, s_{t+1}) = \begin{cases} \dfrac{1}{d_\sigma(\sigma_N, \check{\sigma})} & , \text{ if } t = N-1 \\ 0 & , \text{ else} \end{cases} \quad (23)$$

where $\sigma_N$ is the microstructure associated with the terminal state $s_N$ at the last time step $N$ and where $\check{\sigma}$ denotes the target-microstructure. Under this formulation exactly one non-zero reward signal is emitted per episode on the transition from $s_{N-1}$ to $s_N$. If $\gamma = 1.0$, then the expected future reward $V(s)$ for any state $s \in S$ indicates the expected inverse distance of the associated microstructure $\sigma$ to the target-microstructure $\check{\sigma}$ at the end of the episode.

Under this formulation, non-zero rewards are only occurring at the end of an episode. Especially for long process-paths this can lead to sample-inefficient learning due to the so-called *credit assignment problem* (Sutton and Barto 2018) when using unmodified reinforcement learning algorithms like deep Q-Learning. An efficient way to deal with this problem is *potential-based reward shaping* as introduced in Ng et al. (1999). In order to get a dense reward signal for learning, the original reward function $R$ is substituted by the shaped version

$$R' = R + F, \quad (24)$$

where the shaping function $F$ is in the form

$$F(s_t, s_{t+1}) = \gamma \Phi(s_{t+1}) - \Phi(s_t), \quad (25)$$

and $\Phi : S \rightarrow \mathbb{R}_0^+$ is a potential function of states $s \in S$. As proven in Ng et al. (1999), the optimal policy $\pi^*$ and near optimal policies are invariant to the substitution of $R$ by $R'$ for finite-horizon MDPs with a single terminal state. For finite-horizon MDPs with varying terminal states, the invariance is only guaranteed if $\Phi(s_N) = 0$ for the terminal state $s_N$ in every episode (cf. Grześ 2017; Mannion et al. 2017). During the episode, we use the inverse distance of the associated microstructure to the target-microstructure as potential. The resulting potential function is defined by

$$\Phi(s_t) = \begin{cases} 0 & , \text{ if } t = N \\ \dfrac{1}{d_\sigma(\sigma_t, \check{\sigma})} & , \text{ else.} \end{cases} \quad (26)$$

The shaped rewards function $R'(s_t, a_t, s_{t+1})$ is then given by substituting our original reward function (23) and the shaping function (25) into (24) and then substituting our

potential function $\Phi$ (26):

$$\begin{aligned} R' &= \begin{cases} \dfrac{1}{d_\sigma(\sigma_N, \check{\sigma})} + \gamma \Phi(s_N) - \Phi(s_t) & , \text{ if } t = N-1 \\ \gamma \Phi(s_{t+1}) - \Phi(s_t) & , \text{ else} \end{cases} \\ &= \begin{cases} \dfrac{1}{d_\sigma(\sigma_{t+1}, \check{\sigma})} - \dfrac{1}{d_\sigma(\sigma_t, \check{\sigma})} & , \text{ if } t = N-1 \\ \dfrac{\gamma}{d_\sigma(\sigma_{t+1}, \check{\sigma})} - \dfrac{1}{d_\sigma(\sigma_t, \check{\sigma})} & , \text{ else.} \end{cases} \end{aligned} \quad (27)$$

For undiscounted MDPs, where $\gamma = 1.0$, the cases coincide to $R' = \dfrac{1}{d_\sigma(\sigma_{t+1}, \check{\sigma})} - \dfrac{1}{d_\sigma(\sigma_t, \check{\sigma})}$.

In opposition to $R$, $R'$ emits a dense reward signal for every step during optimization. Due to the policy invariance, $R'$ can be used as a substitute for $R$ without narrowing the *structure-guided processing path optimization* result.

The *single-goal structure-guided processing path optimization* (SG-SGPPO) problem can be solved by a broad range of MDP solving approaches. Surrogate process models of different kinds representing $P(s_{t+1}|s_t, a_t)$ can be estimated from simulation samples as a basis for model-based optimization with dynamic programming methods. This approach would lead to a trade-off between sample-size and accuracy of the surrogate model and hence of the optimization result. In this context, it is important to notice that data acquisition from microstructure evolution simulations is usually computationally expensive. We propose to use model-free reinforcement learning algorithms that directly interact with the simulation model and thereby integrate sampling into the optimization process. Due to its sample-efficiency and the good performance on a broad range of tasks, we regard *deep Q networks* (DQN) as a good basis for *structure-guided processing path optimization*. For an increased sample-efficiency and learning stability, we use *prioritized experience replay* (Schaul et al. 2015b), *double Q-Learning* (Van Hasselt et al. 2016), and *dueling Q-Learning* (Wang et al. 2015) extensions of DQN in our *structure-guided processing path optimization* approaches. Instead of the original reward, expected values of the shaped reward function from (27) are learned. Like in the original DQN paper (Mnih et al. 2015) we choose the initial exploration fraction $\epsilon_0$ to linearly approach $\epsilon_f$ during the first $n_\epsilon$ episodes. The state-space $S$, the action-space $A$, and parameters used for our studies are further specified in Sect. 4.

## Multi-equivalent-goal structure-guided processing path optimization

The multi-equivalent-goal processing path optimization task differs from the single-goal task as depicted in Fig. 4. Instead of a single target-microstructure $\check{\sigma}$, a set $\mathcal{G}$ of equivalent target-microstructures $\check{\sigma}_\mathcal{G}^g \in \mathcal{G}$ is given. The definition

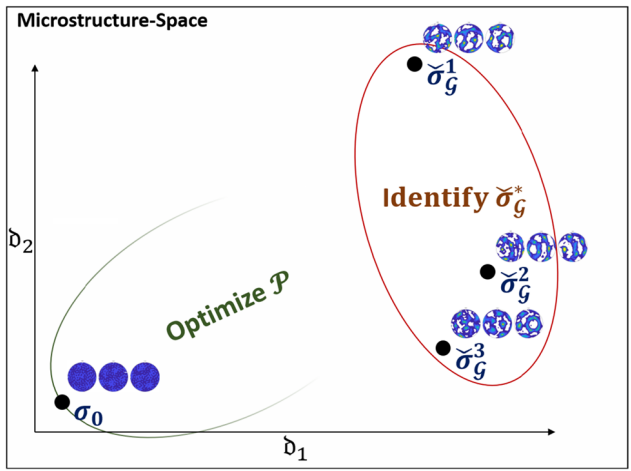

**Fig. 4** Schematic overview of *multi-equivalent-goal structure-guided processing path optimization*. For a set of equivalent target-microstructures $\mathcal{G} = \{\check{\sigma}_{\mathcal{G}}^1, \check{\sigma}_{\mathcal{G}}^2, \check{\sigma}_{\mathcal{G}}^3\}$ the objective is to identify the best reachable target-microstructure $\check{\sigma}_{\mathcal{G}}^*$ from $\mathcal{G}$, while optimizing the processing path $\mathcal{P}$. Here, microstructures are represented by the crystallographic texture, depicted by 100, 110, and 111 pole figure plots

of $\sigma^*$ from (21) then extends to (22). As before $\mathcal{P}^*$ is a processing-path leading from $\sigma_0$ to $\sigma^*$. However, to find $\mathcal{P}^*$, it is now required to also find the best reachable target-microstructure $\check{\sigma}_{\mathcal{G}}^*$.

Because $\check{\sigma}_{\mathcal{G}}^*$ is not known, the reward function from (23) and the shaped reward function from (27) can not be applied directly to the multi-equivalent-goal case. Following (Schaul et al. 2015a), a per-target pseudo reward function $R_g(s_t, a_t, s_{t+1})$ is defined. The per-target shaped reward function $R_g'(s_t, a_t, s_{t+1})$ is derived as described in (27). $R_g$ and $R_g'$ are the multi-target versions of (23) and (27), where $\check{\sigma}$ is replaced by the target-microstructure $\check{\sigma}_{\mathcal{G}}^g \in \mathcal{G}$. Also the optimal policy $\pi_g^*$ now depends on $\check{\sigma}_{\mathcal{G}}^g$. The state-space $S$, the action-space $A$ and the transition probabilities are independent of $\check{\sigma}_{\mathcal{G}}^g$ and remain unchanged.

The basis of our multi-equivalent-goal approach is to define value functions that generalize over targets $\check{\sigma}_{\mathcal{G}}^g$, as proposed for the multi-goal reinforcement learning case in Schaul et al. (2015a). The state value function is thus formed by modifying the state value function $\mathcal{V}_\pi$ from (1) into

$$\mathcal{V}_{g,\pi}(s_t) = \mathbb{E}_{P,\pi}\left[R_g(s_t, a_t, s_{t+1}) + \gamma \mathcal{V}_{g,\pi}(s_{t+1})\right] \quad (28)$$

and the optimal state value function becomes $\mathcal{V}_g^* = \mathcal{V}_{g,\pi_g^*}$. Accordingly, the state-action value function $\mathcal{Q}_\pi$ becomes $\mathcal{Q}_{g,\pi}$ and the *double Q-learning* advantage function $\mathcal{A}_\pi$ is $\mathcal{A}_{g,\pi}$, with $\mathcal{Q}_g^* = \mathcal{Q}_{g,\pi_g^*}$ and $\mathcal{A}_g^* = \mathcal{A}_{g,\pi_g^*}$.

As done in the single-goal approach, we use the extended *deep Q-Learning* algorithm for value-function approximation. The generalization of the estimated values across multiple targets is enabled by generalizing the single-goal function approximators during the approximation of the target dependent value functions. The optimal Q-function $\mathcal{Q}_g^{\prime*}$ is approximated by $\mathcal{Q}'(s, a, \mathfrak{d}(\check{\sigma}_{\mathcal{G}}^g), \theta)$ with model parameters $\theta$, where $\mathfrak{d}(\check{\sigma}_{\mathcal{G}}^g)$ returns a feature vector of the target-microstructure $\check{\sigma}_{\mathcal{G}}^g$. The extensions described in Sect. 2.2 can be applied directly to the generalized *deep Q-learning* approach. When using *dueling Q-learning*, $\mathcal{V}_g^{\prime*}$ is approximated by $\mathcal{V}'(s, \mathfrak{d}(\check{\sigma}_{\mathcal{G}}^g), \theta)$, and $\mathcal{A}_g^{\prime*}$ by $\mathcal{A}'(s, a, \mathfrak{d}(\check{\sigma}_{\mathcal{G}}^g), \theta)$ accordingly.

As described above, the major aim of the proposed multi-equivalent-goal approach is to prioritize promising targets from $\mathcal{G}$ during optimization and thereby enable a tractable processing path optimization. We use the expected distance of the terminal microstructure $\sigma_N$ to the target-microstructure $\check{\sigma}_{\mathcal{G}}^g$ as criterion for target-prioritization at the beginning of each episode. The state value $\mathcal{V}_g^*(s_0)$ of the original reward formulation (23) can be used to prioritize targets $\check{\sigma}_{\mathcal{G}}^g$ because it reflects the inverse of the expected distance of $\sigma_N$ to $\check{\sigma}_{\mathcal{G}}^g$. However, due to reward shaping, the optimal state value for the shaped reward function $\mathcal{V}_g^{\prime*}(s)$ is learned instead. When $\Phi(s_N) = 0$ for every terminal state $s_N$, following (Ng et al. 1999), the relation between $\mathcal{V}_g^*(s)$ and $\mathcal{V}_g^{\prime*}(s)$ is

$$\mathcal{V}_g^{\prime*}(s) = \mathcal{V}_g^*(s) - \Phi(s). \quad (29)$$

Thus, we can reconstruct $\mathcal{V}_g^*(s_0)$ by

$$\mathcal{V}_g^*(s_0) = \mathcal{V}_g^{\prime*}(s_0) + \frac{1}{d_\sigma(\sigma_0, \check{\sigma}_{\mathcal{G}}^g)} \quad (30)$$

and, on this basis, estimate the best reachable target-microstructure by

$$\check{\sigma}_{\widetilde{\mathcal{G}}} = \arg\max_{\check{\sigma}_{\mathcal{G}}^g \in \mathcal{G}}\left[\mathcal{V}'(s_0, \mathfrak{d}(\check{\sigma}_{\mathcal{G}}^g), \theta) + \frac{1}{d_\sigma(\sigma_0, \check{\sigma}_{\mathcal{G}}^g)}\right], \quad (31)$$

where $\mathcal{V}'$ is the approximated state value for the shaped rewards and $\mathfrak{d}(\check{\sigma}_{\mathcal{G}}^g)$ is the description of $\check{\sigma}_{\mathcal{G}}^g$.

As for the choice of processing actions during learning, the proposed learning agent decides for a targeted $\check{\sigma}_{\mathcal{G}}$ per episode in an $\epsilon$-greedy manner. Per episode, the target-microstructure $\check{\sigma}_{\mathcal{G}} \in \mathcal{G}$ is chosen by

$$\check{\sigma}_{\mathcal{G}} = \begin{cases} \check{\sigma}_{\widetilde{\mathcal{G}}} & \text{, if } u > \epsilon^G \\ \text{random}(\mathcal{G}) & \text{, else,} \end{cases} \quad (32)$$

where $u \sim \mathcal{U}(0, 1)$ is randomly drawn from the uniform distribution over $[0, 1]$ and random$(\mathcal{G})$ returns a random sample from $\mathcal{G}$. The goal selection exploration rate $\epsilon^G$ is decoupled from the action selection exploration rate $\epsilon$. The initial $\epsilon_0^G$ linearly approaches $\epsilon_f^G$ during the first $n_{\epsilon^G}$ episodes.

Inspired by the multi-goal augmentation in *hindsight experience replay* (Andrychowicz et al. 2017), we add experience-tuples $(s_t, a, R_{\mathfrak{g}}, s_{t+1}, \mathfrak{d}(\breve{\sigma}_{\mathcal{G}}^g))$ for all targets from $\mathcal{G}$ with the associated potential reward to the replay memory $\mathcal{D}$, independently of the current target-microstructure. Thereby, we further increase the sample-efficiency of our multi-equivalent-goal approach.

The overall multi-equivalent-goal approach is outlined in Listing 1, where $s$ and $s'$ are short forms of $s_t$ and $s_{t+1}$. At the beginning of each episode, the episode target $\breve{\sigma}_{\mathcal{G}}$ is picked in an $\epsilon^{\text{G}}$-greedy manner (lines 2 and 3). For the sake of simplicity, we condense the per-episode execution of *multi-goal DQN* in line 4. By *multi-goal DQN* we denote the DQN algorithm combined with models that generalize over target-microstructure descriptions $\mathfrak{d}(\breve{\sigma}_{\mathcal{G}})$ as described above. The replay memory $\mathcal{D}$ is not updated during the episode. Instead, the set of experience tuples from the current episode $\mathcal{D}_e$ are returned by DQN. The replay memory $\mathcal{D}$ is then updated by hypothetical experiences for all goals from $\mathcal{G}$ (lines 5, 6, 9).

MEG-SGPPO($\#episodes, \mathcal{G}, \epsilon_0^{\text{G}}, \epsilon_1^{\text{G}}, n_\epsilon^{\text{G}}, DQNparameters$)

```
1  for e = 1 to #episodes
2      ε^G ← max (ε_1^G, ε_0^G − e/n_{ε,G} (ε_0^G − ε_1^G))
3      σ̆_G ← Eqs. (31), (32)
4      D_e ← execute multi-goal DQN w. σ̆_G for one episode
5      for σ̆_G^g ∈ G
6          D_e^g ← {(s,a,s',R, 𝔡(σ̆_G^g))|(s,a,s',R) ∈ D_e}
7          for (s,a,s', 𝔤̃) ∈ D_e^g
8              R ← R_g' (Eq. 27), where σ̃ = σ̆_G^g
9          D ← D ∪ D_e^g
```

Listing 1. Multi-equivalent-goal structure-guided processing path optimization (MEG-SGPPO).

# Application to crystallographic texture evolution

## ODF distance

For the reward functions defined in the previous section, a distance measure in microstructure space is required. In this contribution, we limit ourselves to crystallographic texture as the only microstructural feature to describe microstructures $\sigma$. To enable distance measurements between ODFs, we propose an orientation-histogram representation, achieved by ODF binning in orientation-space. For the proposed histogram approach, a set of approximately equally sized bins in orientation space is required. Due to crystal symmetries, the orientation space can be reduced to a crystal symmetry dependent *fundamental zone*.

For a set of $J$ orientations $o_j \in O_J^\Omega$, which are assumed to be equally distributed in the *fundamental zone* of a given crystal system $\Omega$ and an $SO(3)$ distance metric $\phi : SO(3) \times$ $SO(3) \to \mathbb{R}_0^+$, we propose a nearest-neighbor bin assignment as is described in the following.

For a single crystal orientation $h$ the assignment vector $\mathbf{w}_h$ is defined by

$$w_{h,j} = \begin{cases} 1 & \text{, if } o_j = \arg\min_{o \in O_J^\Omega}[\phi_\Omega(o, h)] \\ 0 & \text{, else,} \end{cases} \quad (33)$$

where $w_{h,j}$ is the j-th vector component of $\mathbf{w}_h$ and $\phi_\Omega :$ $SO(3) \times SO(3) \to \mathbb{R}_0^+$ denotes the minimal distance $\phi$ for all equivalent orientations regarding $\Omega$. Hence,

$$\phi_\Omega(o, h) = \min_{(\bar{o}, \bar{h}) \in \Psi_\Omega(o) \times \Psi_\Omega(h)} \phi(\bar{o}, \bar{h}), \quad (34)$$

where $\Psi_\Omega(h)$ is the set of all equivalent orientations of the orientation $h$ for the crystal system $\Omega$.

For a representative sample set $H$ of orientations $h$ and associated volume $V(h)$ from the crystallographic texture $\sigma$, the orientation histogram of $\sigma$ is defined by the vector

$$\mathbf{b}_\sigma = \frac{1}{V} \sum_{h \in H} V(h) \cdot \mathbf{w}_h. \quad (35)$$

For any two crystallographic textures $\sigma_a$, $\sigma_b$, we then use the Chi-Square distance of the associated orientation histogram representations

$$\chi^2(\mathbf{b}_{\sigma_a}, \mathbf{b}_{\sigma_b}) = \sum_{j=0}^{J} \frac{(\mathbf{b}_{\sigma_a, j} - \mathbf{b}_{\sigma_b, j})^2}{(\mathbf{b}_{\sigma_a, j} + \mathbf{b}_{\sigma_b, j})} \quad (36)$$

as the distance function $d_\sigma(\sigma_a, \sigma_b)$.

We use $\phi(q_1, q_2) = \min(||q_1 - q_2||, ||q_1 + q_2||)$ as basis distance function, where $q_1, q_2$ are orientations encoded as unit quaternions. $\phi$ is a metric in $SO(3)$ (Huynh 2009). To generate the set of approximately uniformly distributed orientations $O_J^\Omega$, we use the optimization based approach of Quey et al. Quey et al. (2018). For efficient nearest-neighbor queries, we use the *k-d tree* algorithm.

To smoothen the distance function, in practice we use a soft assignment $\tilde{\mathbf{w}}_h$ instead of the hard assigned weights $\mathbf{w}_h$ from (33). $\tilde{\mathbf{w}}_h$ is defined by

$$\tilde{w}_{h,i} = \begin{cases} \dfrac{\phi_\Omega(o_i, h)}{\sum_{o_j \in N} \phi_\Omega(o_j, h)} & \text{, if } o_i \in \mathcal{N} \\ 0 & \text{, else,} \end{cases} \quad (37)$$

where $\mathcal{N} = \text{NN}_k(O_J^\Omega, h, \phi_\Omega)$ is the set of the $k$ nearest neighbors of $h$ from $O_J^\Omega$. We study the effect of the parameters $J$ and $k$ on the ODF distance function and regarding the representation quality in Sect. 5.

## Application scenario

To evaluate the proposed algorithms, a texture generating process based on the Taylor-type material model is set up. The process consists of subsequent steps of uniaxial tension or compression, variably oriented to the reference coordinate system of the material model. In each step, a deformation $\hat{F}$ is applied:

$$\hat{F} = R\bar{F}R^\top, \tag{38}$$

with a rotation defined by the rotation matrix $R$ and

$$\bar{F} = \begin{bmatrix} \tilde{F}_{11} & 0 & 0 \\ 0 & F_{22} & 0 \\ 0 & 0 & F_{33} \end{bmatrix}. \tag{39}$$

While $\tilde{F}_{11}$ is variable, $F_{22}$ and $F_{33}$ are adjusted by the simulation framework until the stresses are in balance.

The process parameters that have to be optimized are

1. the magnitude of $\tilde{F}_{11}$
2. the rotation matrix $R$

After each process step, the reorientation of the single crystals is calculated based on the rigid body rotation $R_e$, which follows from the polar decomposition of the elastic part of the deformation gradient $F_e$ (Ling et al. 2005). To avoid unrealistically large deformations, the process is limited to induce an equivalent strain of 70%. Based on the reached crystallographic textures, material properties can be calculated such as the Young's modulus $E_{ii}$ in $ii$-direction regarding the reference coordinate system. However, we want to remark here, that when using the elastic constants for iron single crystals by neglecting the alloying elements in steel (Eq. (10)), the Young's modulus is slightly overestimated (Hoffmann 2010).

For the evaluation studies, a processing path consists of up to $K$ subsequent process-steps. The per-step process control action $a_t = (f_t, \mathbf{q}_t)$ is composed of the deformation magnitude $f_t \in [-1.0, 1.0]$, representing $\tilde{F}_{11}$ at process-step $t$ and the rotation $R$ at processing-step $t$ encoded as unit quaternion $\mathbf{q}_t \in \{x \in \mathbb{H} \,|\, ||x||_2 = 1\}$. In our evaluation scenario, we restrict $f_t$ to the interval $[-0.02, 0.02]$. For our value based deep reinforcement learning approaches, we discretize the action space $A$ by setting $f \in \{0.02, -0, 02\}$ and $\mathbf{q} \in O_{100}$, where $O_{100}$ is a set of 100 nearly uniformly distributed unit quaternions. Furthermore, the action space contains a *no-op* action to be used when no further improvement regarding the distance to the target-microstructure during the episode is expected.

The state-space $S$ consists of descriptions of reachable microstructures. As microstructures $\sigma$ are described by the according ODF, states $s \in S$ are represented by ODF descriptors. For this purpose, we use *generalized spherical harmonics* coefficients $\zeta_l^\Omega$, as introduced in Sect. 2.4. We truncate the series at $l = 8$. For the used material with cubic crystal symmetry, the ODF is represented by 21 independent complex-valued coefficients. In addition to the GSH microstructure description, the state vector $s$ contains the current time-step and equivalent strain.

## Implementation details

Single-goal evaluation optimization runs consist out of 100 subsequent process episodes with a maximal length of $K = 100$ each. The initial microstructure state is a uniform orientation distribution (often also called grey texture) and consists of 250 equally weighted orientations which are nearly uniformly distributed in the *fundamental zone* of the cubic crystal symmetry. If not stated otherwise for specific results, experiments are conducted with the following hyperparameters:

– $d_\sigma$ as defined in (36) with histogram parameters $J = 512$ and $k = 3$.
– A discount factor of $\gamma = 1.0$.
– Deep Q-learning as described in Sect. 2.2 as basic algorithm where the target-network is updated every $n_\theta = 250$ time-steps.
– Prioritized replay as described in Sect. 2.2, with $\alpha_{PER} = 0.6$ and $\beta_0 = 0.4$.
– Double-Q-learning and dueling-Q-learning as described in Sect. 2.2.
– An $\epsilon$-greedy policy, with an initial exploration rate $\epsilon_0 = 0.5$ and the final exploration-rate $\epsilon_f = 0.1$, with $n_\epsilon = 50$.
– Q-networks with hidden layer sizes of [128, 64, 32], layer normalization and *ReLU* activation functions. The learning process starts after 100 control-steps. The networks are trained after each control-step with a mini-batch of size 32.
– ADAM is used as optimizer for neural network training, with a learning rate of $5e^{-4}$.

Multi-equivalent-goal optimization runs consist of 200 subsequent process episodes. Hyperparameters differ from the single-goal experiments in the following points:

– Due to the higher data complexity we expanded the neural network to 4 hidden layers of sizes [128, 256, 256, 128].
– Per time-step the network is trained with four mini-batches instead of one.
– For improved stability, we reduced the target-network update frequency to $n_\theta = 500$ steps.
– Exploration parameters are $\epsilon_0 = 0.5$ and the final exploration-rate $\epsilon_f = 0.0$, with $n_\epsilon = 190$. The $\epsilon_G$-greedy

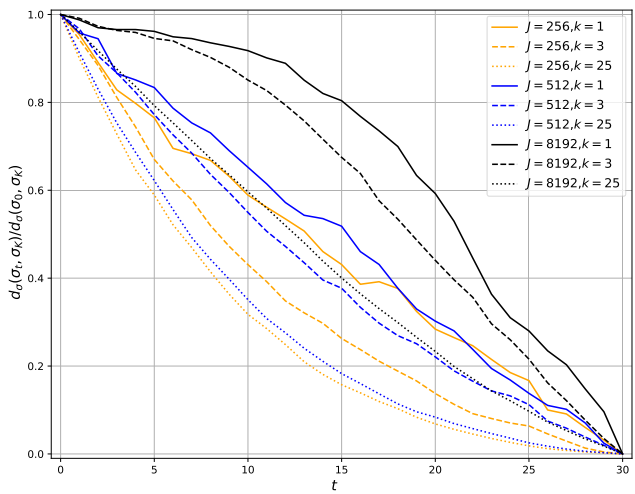

**Fig. 5** Comparison of the effect of orientation histogram parameters on the resulting ODF distance function. Relative distance $d_\sigma(\sigma_t, \sigma_K)/d_\sigma(\sigma_0, \sigma_K)$ over $t$ for an exemplary process with static process parameters, where $\sigma_t$ is the microstructure at time-step $t$ and $\sigma_K$ is the microstructure resulting from the

**Table 2** Mean absolute error of Young's modulus $E$ in 11, 22 and 33 direction for various orientation histogram parameters $J$, $k$ in MPa

|  |  | $k$ | | |
|---|---|---|---|---|
|  |  | 1 | 3 | 25 |
|  | 256 | 574 | 238 | 401 |
| $J$ | 512 | 493 | 201 | 275 |
|  | 8192 | 196 | 77 | 57 |

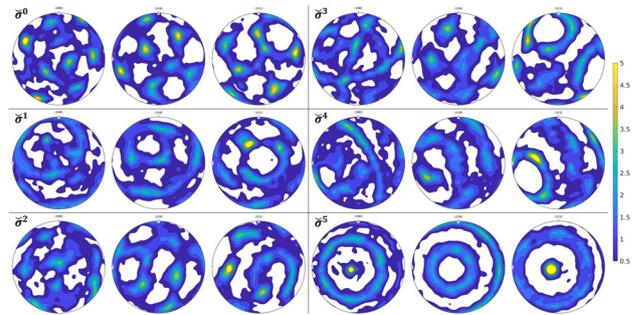

**Fig. 6** Pole figure triples of the target-textures from $G$, used throughout the single-goal studies. Miller indices of the pole figures from left to right: [100, 110, 111]

target choice is parametrized with $\epsilon_0^G = 1.0$, $\epsilon_f^G = 0.0$, $n_{\epsilon^G} = 190$.

The Tailor-type material model and the simulation framework are implemented in Fortran. The reinforcement learning code is implemented in python. The reinforcement learning environment code supports the *openAI gym* interface (Brockman et al. 2016). Reinforcement learning agents are implemented based on *stable baselines*, an open-source collection of implementations of deep reinforcement learning algorithms. The *sciPy k-d tree* implementation is used for orientation distribution representation calculations (Virtanen et al. 2020). *Neper* (Quey et al. 2011) is used to generate the approximately uniformly distributed orientations following the approach from Quey et al. (2018). Approximately uniformly distributed orientations are used to define the bins of the orientation histogram described in Sect. 3, the initial texture for our experiments, and the discrete action-space as described in this Section. The pole figures in the figures of Sect. 5 are created with MTEX (Bachmann et al. 2010). Experiments have been conducted on workstations with 20 2.2 GHz CPU cores (mainly used by the simulation) and GTX 1080 Ti GPUs (for artificial neural network training).

## Results and discussion

### Orientation histogram parameters

The ODF distance is represented by the distance between orientation histograms as presented in Sect. 4. A. The effect of the histogram formation parameters $J$ (number of his-togram bins) and $k$ (soft assigment neighborhood size) on the ODF distance function is shown in Fig. 5. For various $(J, k)$-settings and an exemplary process path of the distance to the crystallographic texture $\sigma_T$ in relation to the initial distance $d_\sigma(\sigma_t, \sigma_T)/d(\sigma_0, \sigma_T)$ is plotted over time steps $t$. The exemplary process path is sampled from the process described in Sect. 4.2 by starting from a uniform ODF and applying a constant deformation in direction $q = (1, 0, 0, 0)$ for 30 process steps. When constantly approaching $\sigma_T$ as in the exemplary process path, a monotone decreasing distance function is assumed in the reward function formulations ((23) and (27)). For $k = 1$ (solid lines) the distance-trajectory is not strictly monotone. A clear smoothing effect, depending on the soft assignment parameter $k$, can be observed.

The information loss induced by the orientation histogram representation of the ODF can be seen from the effect on the material properties. Table 2 lists the mean error of Young's modulus induced by the discretization and smoothing for various parameters $(J, k)$. The mean absolute errors are calculated based on 1000 crystallographic textures which are randomly sampled from a set of textures reachable by the process and by calculating Young's modulus for the original texture and the orientation histogram of the texture. While increasing the number of bins, $J$ has a clear and monotone positive impact regarding the mean absolute error in both cases. An adequate soft-assignment parameter for the sampled crystallographic textures in our search-grid is $k = 3$. The computational cost of the nearest-neighbor search is highly effected by $J$. A reasonable hyperparameter setting, which we use throughout our studies, is $(J = 512, k = 3)$.

**Table 3** Young's modulus of target textures from $G$ in GPa

| Goal | $E_{11}$ | $E_{22}$ | $E_{33}$ |
|---|---|---|---|
| $\check{\sigma}^0$ | 221 | 223 | 221 |
| $\check{\sigma}^1$ | 216 | 221 | 212 |
| $\check{\sigma}^2$ | 223 | 219 | 214 |
| $\check{\sigma}^3$ | 222 | 218 | 223 |
| $\check{\sigma}^4$ | 224 | 218 | 219 |
| $\check{\sigma}^5$ | 227 | 226 | 233 |

## Single-goal processing path optimization

A set of six target textures is used throughout the evaluation of the single-goal algorithm: $G = \{\check{\sigma}^0, \check{\sigma}^1, \check{\sigma}^2, \check{\sigma}^3, \check{\sigma}^4, \check{\sigma}^5\}$. The textures are visualized in Fig. 6 in the form of pole figures. The calculated properties for the sampled target-textures are listed in Table 3. The six target textures have been randomly sampled from a set of textures, which are likely reachable by the process. The following criteria have been applied to guarantee the diversity of the sample:

1. $d_\sigma(\check{\sigma}^i, \check{\sigma}^j) > 1.2$ for all $\check{\sigma}^i, \check{\sigma}^j$ in $G$.
2. $\min_{\check{\sigma} \in G} d_\sigma(\check{\sigma}, \hat{\sigma}) > 0.75$, where $\hat{\sigma}$ is the uniform ODF.

The single-goal algorithm is applied to each target-texture from $G$ separately. In the remainder of this chapter, we use the more general term microstructure instead of the specific term texture to be consistent with the theoretical and methodical chapters. The solution variety (due to stochastic exploration and function approximation) has been covered by multiple independent optimization runs of 100 reinforcement learning episodes each. The learning objective is to find a processing path from the initial microstructure close to the target-microstructure $\check{\sigma}$.

During learning, each episode ends after $K$ time-steps or when the equivalent-strain exceeds its limit. Processing paths can be pruned in hindsight. The optimal sub-path of episode $e$ regarding the optimization task is the path from the initial microstructure $\sigma_0$ to $\tilde{\sigma}_e = \arg\min_{\sigma \in \mathcal{S}_e} d_\sigma(\sigma, \check{\sigma})$, where $\mathcal{S}_e \subset S$ is the set of microstructures on the processing path of episode $e$. The mean distance of $\tilde{\sigma}_e$ to the according target-microstructure per episode is plotted for three independent optimization runs per target-microstructure from $G$ on the left of Fig. 7 together with the per-target 95% confidence intervals. On the right of Fig. 7 the goal-distance of the current best processing path is plotted over episodes.

Qualitative results of a single exemplary optimization run are shown on the left of Fig. 8. On the right of Fig. 8, the results for a single episode are shown. The visualized results belong to the best of the three independent optimization runs with the target-microstructure $\check{\sigma}^0$. On the left, the goal-distance of the current best processing path is plotted for the single optimization run as in Fig. 7 with pole figures of $\tilde{\sigma}_0, \tilde{\sigma}_4, \tilde{\sigma}_{10}, \tilde{\sigma}_{21}, \tilde{\sigma}_{40}$ and the resulting microstructure of the best found processing path $\tilde{\sigma}_{97}$. On the upper right, the associated target-microstructure $\check{\sigma}^0$ is visualized. On the right of Fig. 8 the processing path of episode 97 is plotted in the form of a scatter plot, where the control action displacement direction (vertical axis) is plotted along with the displacement sign (color) against the processing time-step number. Additionally, the distance of the microstructures within the episode $\sigma_t$ to the target-microstructure $\check{\sigma}^0$ is plotted as a line plot. Each dot in the scatter plot represents an action $a_t \in A$, where the positions on the y-axis mark the chosen load orientation $q \in O_{100}$ and the colors mark the chosen load. The dashed vertical line marks the time-step of the best microstructure of the episode $\tilde{\sigma}_{97}$ in terms of the distance to the target-microstructure. In this case, the solid vertical line marks the end of the episode due to the equivalent load criterion. Figure 9 shows the best found processing path of all optimization runs for all target-microstructures from $G$ in the top row and pole figures of the corresponding processing path results in the middle row, and below the processing path result of the worst optimization run for the corresponding target-microstructure. Figure 9 in combination with Fig. 7 shows, that for some microstructures ($\check{\sigma}^3, \check{\sigma}^5$) the processing paths are composed of only very few different actions, which are found in the first few episodes.

Figure 10 shows ablation study results on target-microstructure $\check{\sigma}^0$ for the single-goal algorithm. Per setting the mean and 95% confidence interval is plotted for three independent optimization runs. The plot shows that the addition of the three DQN extensions, namely *Prioritized Experience Replay*, *Double Q-Learning*, *Dueling Q-Learning*, has no major impact on the optimization outcome but slightly stabilizes and accelerates the convergence. The results of the single-goal algorithm with the pure reward signal from (23) (green curve) reveals the impact of reward shaping on the convergence speed, and hence on the data efficiency of the algorithm.

## Multi-equivalent-goal processing path optimization

According to the results reported in the previous subsection, especially for $\check{\sigma}^2$ and $\check{\sigma}^4$ no satisfactory processing paths were found by the single-goal algorithm. In contrast to the single-goal approach, the multi-equivalent-goal optimization approach aims to reach one of several target-microstructures that are equivalent with respect to the material properties. In this subsection, we report and discuss the results of the *multi-equivalent-goal structure-guided processing path optimization* (MEG-SGPPO) algorithm as introduced in Sect. 3.3. To study the advantages of the multi-equivalent-goal approach we sampled target-microstructure sets $\mathcal{G}^{\sigma\,4\text{-equiv}}$ and $\mathcal{G}^{\sigma\,2\text{-equiv}}$, where

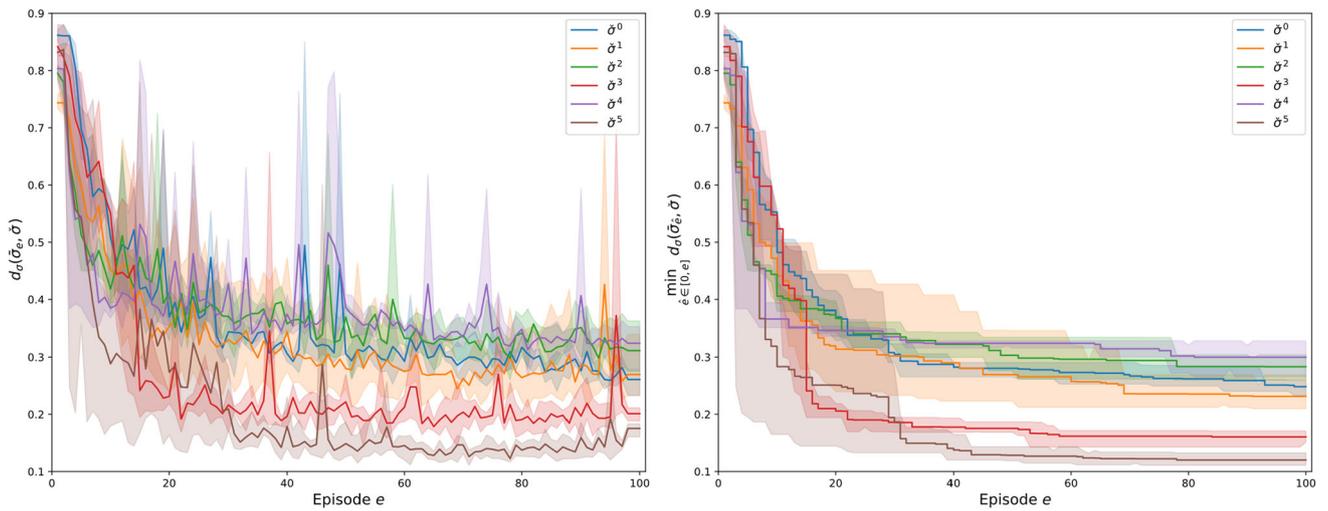

**Fig. 7** Single-goal reinforcement learning results for the target-microstructures $G$. (left) The mean and the 95% confidence interval of the goal-distance of $\check{\sigma}$ per episode $e$ and $\check{\sigma} \in G$ (color). (right) Shortest previous goal-distance during the optimization run

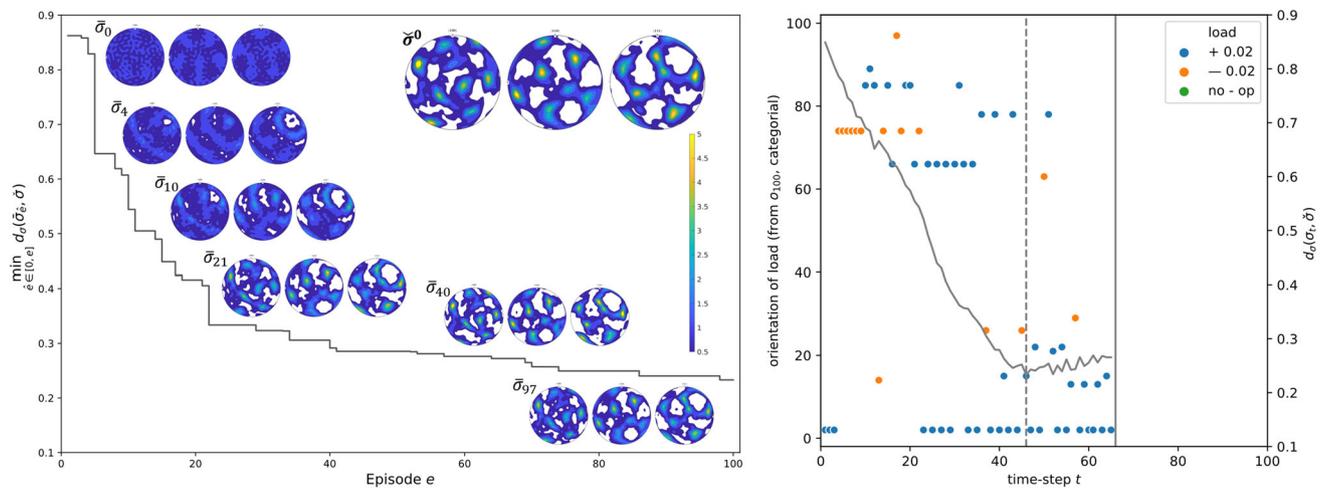

**Fig. 8** Single-goal optimization. (left) Best optimization run for $\check{\sigma}^0 \in G$ with pole figure plots. Miller indices of the pole figures from left to right: [100, 110, 111]. (right) Episode 97 of the exemplary optimization run with control actions and the goal-distance visualized per time-step

- $\mathcal{G}^{\sigma 4\text{-equiv}}$ consists of 10 distinctive $\check{\sigma}^4$-equivalent microstructures $\check{\sigma}^g_{4\text{-equiv}}$ and
- $\mathcal{G}^{\sigma 2\text{-equiv}}$ consists of 10 distinctive $\check{\sigma}^2$-equivalent microstructures $\check{\sigma}^g_{2\text{-equiv}}$.

For this purpose, we defined a microstructure $\check{\sigma}^g_{k\text{-equiv}}$ to be equivalent to $\check{\sigma}^k \in G$ if the Young's moduli $E_{jj}(\check{\sigma}^g_{k\text{-equiv}})$ of $\check{\sigma}^g_{k\text{-equiv}}$ are within the range of $+/- 0.5$ GPa of $E_{jj}(\check{\sigma}^k)$ for $j \in 1, 2, 3$.

The results of MEG-SGPPO on $\mathcal{G}^{\sigma 4\text{-equiv}}$ and $\mathcal{G}^{\sigma 2\text{-equiv}}$ are visualized in Fig. 11. In both cases, we conducted one optimization run and present quantitative results per-episode and qualitative results for the best episodes in the bottom row. As in Fig. 8 (left), the distance of nearest microstructure $\bar{\sigma}_e$ to the actual chosen target-microstructure from the set $\mathcal{G}^{\sigma\text{-equiv}}$ is plotted per episode $e$. The per-episode target-microstructure choice of the algorithm is depicted by the color of the respective scatter dot. The type of the specific selection (explorative/greedy) is represented by the shape of the respective scatter dot. Due to the decreasing goal-picking exploration factor $\epsilon^G$, the variety of the targeted microstructures is high in the beginning and then decreases during the learning process.

As the scatter plot for $\mathcal{G}^{\sigma 4\text{-equiv}}$ (Fig. 11 (left)) and the underlying pole figure plots of $\check{\sigma}^1_{4\text{-equiv}}$ suggest, this target-microstructure is closer to the grey texture than the other equivalent targets from $\mathcal{G}^{\sigma 4\text{-equiv}}$. Although the algorithm finds good processing paths for other target-microstructures ($\check{\sigma}^0_{4\text{-equiv}}$ and $\check{\sigma}^8_{4\text{-equiv}}$), it rapidly commits to $\check{\sigma}^1_{4\text{-equiv}}$ due to the low distance $d_\sigma(\sigma_0, \check{\sigma}^1_{4\text{-equiv}})$, and does not change this target

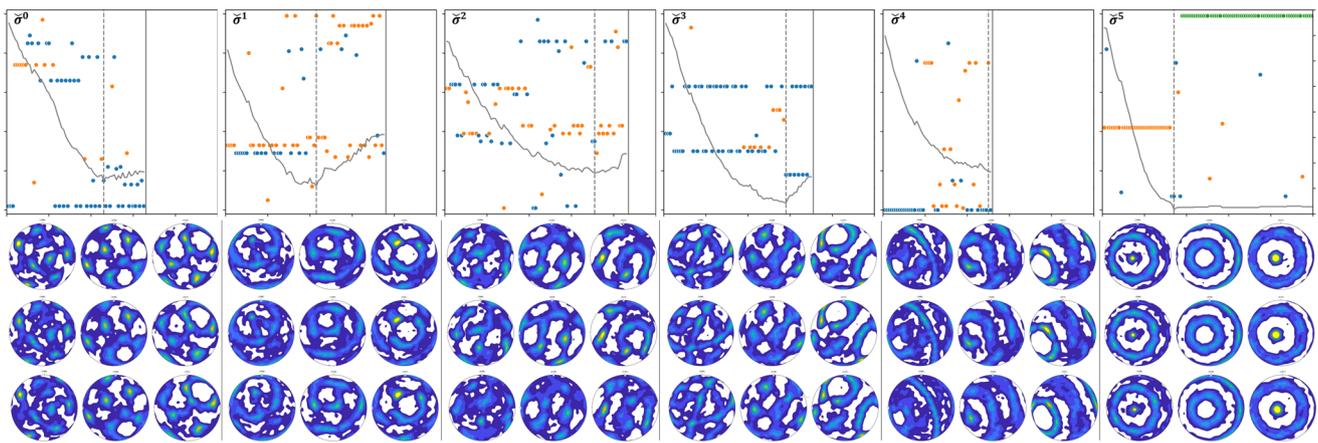

**Fig. 9** Single-goal reinforcement learning results per target-microstructure $\check{\sigma} \in G$. (top) Best found processing path per target-microstructure (plot axis parameters are identical to Fig. 8, right). (bottom) Three lines of per-target [100, 110, 111] pole figures. First line: target-textures from $G$. Second line: resulting textures of the best found processing paths in the best of the three independent optimization runs. Third line: resulting textures of the best found processing paths in the worst of the three independent optimization runs

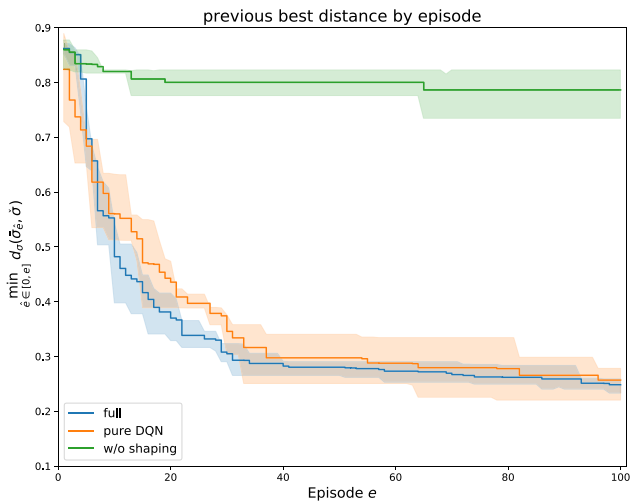

**Fig. 10** Ablation results (mean and 95% confidence interval) for $\check{\sigma}^0$. Best previous goal distance by episode for three independent optimization runs each of various settings. Blue: results of the full featured algorithm as reported in Fig. 7, orange: pure DQN (without *Prioritized Experience Replay*, *Double Q-Learning*, *Dueling Q-Learning*), green: results obtained without reward-shaping

in the greedy episodes during the experiment. In the case of $\mathcal{G}^{\sigma 2\text{-equiv}}$ (Fig. 11 (right)), the algorithm tries to find a good processing path for $\check{\sigma}^3_{2\text{-equiv}}$ during the first 105 episodes. After obtaining better results for $\check{\sigma}^3_{2\text{-equiv}}$ in episode 103 the greedy target fluctuates between $\check{\sigma}^3_{2\text{-equiv}}$ and $\check{\sigma}^6_{2\text{-equiv}}$. After episode 144 the algorithm finally decides for $\check{\sigma}^3_{2\text{-equiv}}$ due to the repeating better results for $\check{\sigma}^3_{2\text{-equiv}}$ after episode 103 compared to $\check{\sigma}^6_{2\text{-equiv}}$.

The best result obtained during the single-goal approach experiments for the respective original microstructure (0.2656 for $\check{\sigma}^4$, 0.2601 for $\check{\sigma}^2$) is depicted as a baseline by the grey dashed line in Fig. 11, top-left and top-right. This baseline is exceeded by MEG-SGPPO during the first 25 episodes for $\mathcal{G}^{\sigma 4\text{-equiv}}$ and during the first 75 episodes for $\mathcal{G}^{\sigma 2\text{-equiv}}$. In both cases, the algorithm yields notably better results during the 200 episodes than the single-goal approach does on the original target-microstructure.

On the bottom of Fig. 11, qualitative results are plotted for the best episodes of the best reached target-microstructure ($\check{\sigma}^1_{4\text{-equiv}}$ in Fig. 11 (left), and $\check{\sigma}^3_{2\text{-equiv}}$ in Fig. 11 (right)) and the second best reached target-microstructure ($\check{\sigma}^0_{4\text{-equiv}}$ in Fig. 11 (left) and $\check{\sigma}^6_{2\text{-equiv}}$ in Fig. 11 (right)). For each of the four qualitative results, we show on the left the pole figures of the target texture and below the pole figures of the resulting texture of the best found processing path $\bar{\sigma}_e$. Right of the pole figures, the best found processing path is shown in form of a scatter plot of the chosen action per processing step and - overlaid as a line chart - the corresponding microstructure distance to the respective target-microstructure. The pole figures give an impression of the similarity between the results and the respective target-structures. The according process diagrams show the complexity of the found processes while approaching the process targets.

To investigate the benefit of the proposed multi-goal augmentation, we conducted repeated optimization runs with and without the augmentation of the replay memory by data for non-pursued goals. Results are depicted in Fig. 12 in the form of the best previous result plot known from the last subsection. The basis of the plot are five independent optimization runs with the full-featured MEG-SGGPO algorithm (blue)

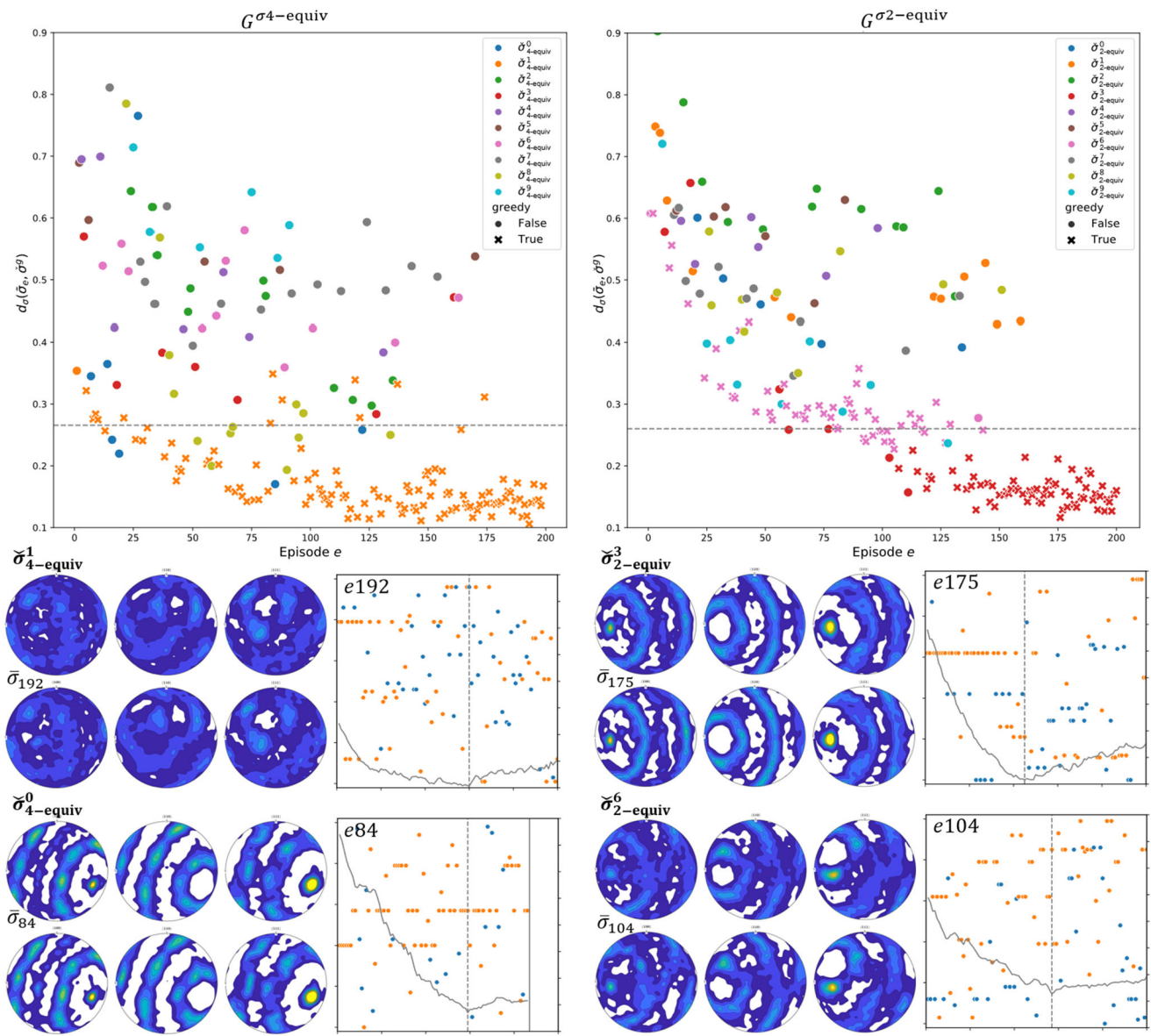

**Fig. 11** Results of the multi-equivalent-goal algorithm for a set of 10 microstructures $\check{\sigma}^g_{4\text{-equiv}} \in \mathcal{G}^{\sigma 4\text{-equiv}}$ equivalent to $\check{\sigma}^4 \in G$ (left), and 10 microstructures $\check{\sigma}^g_{2\text{-equiv}} \in \mathcal{G}^{\sigma 2\text{-equiv}}$ equivalent to $\check{\sigma}^2 \in G$ (right). Top: Scatter-plots of the best distance to the selected target-microstructure $d_\sigma(\bar{\sigma}_e, \check{\sigma}^g_{k\text{-equiv}})$ over episode numbers, where the dot color denotes the selected target-microstructure per episode and the dot shape the type of the target-microstructure selection. The best reached goal-distance to the respective original target-microstructure from the single-goal exper-

iments (0.2656 for $\check{\sigma}^4 \in G$ and 0.2601 for $\check{\sigma}^2 \in G$) is marked by a grey dashed line. Bottom: Qualitative results of the two microstructures closest to one of the equivalent target-microstructures at the corresponding episode e for each of the two experiments: Pole figures of the texture of the chosen chosen target $\check{\sigma}^g_{k\text{-equiv}}$ and the nearest reached texture $\bar{\sigma}_e$ with its episode number e (projections 100, 110, 111, where the color map is identical to Fig. 6). To the right of the pole figures the processing path of the episode e, according to $\bar{\sigma}_e$ is plotted. The plot axis parameters are identical to Fig. 8 (right) and five independent optimization runs without augmentation (orange) on the target-microstrurecture set $\mathcal{G}^{\sigma 2\text{-equiv}}$. As the plot suggests, the augmentation improves the overall optimization result.

## Conclusion and outlook

In the present paper, two deep reinforcement learning approaches are proposed to solve processing path optimization problems with the objective function and state representation in the structure space. The *single-goal structure-guided processing path optimization* (SG-SGPPO) algorithm

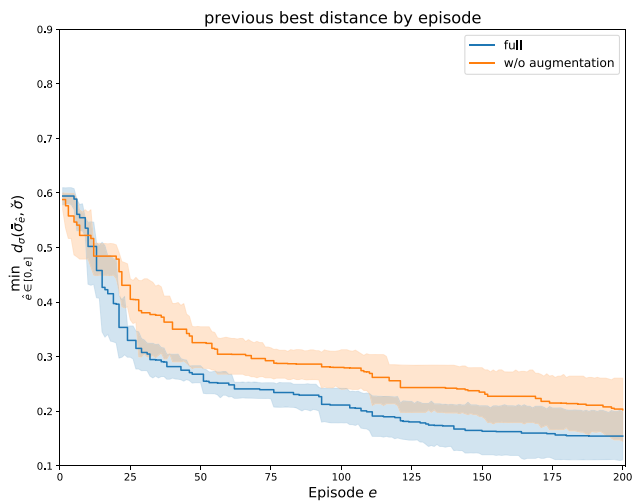

previous best distance by episode

**Fig. 12** Ablation results (mean and 95% confidence interval) for the target-microstructure set $\mathcal{G}^{\sigma 2\text{-equiv}}$. Best previous goal distance for episode for five independent optimization runs per setting. Blue: full featured approach, as introduced and used for other MEG-SGGPO results reported. Orange: without multi-goal augmentation

combines function-based reinforcement learning techniques and a reward shaping mechanism for *structure-guided processing path optimization*. However, from the materials design point of view, typically various microstructural compositions are equivalent with respect to a desired macroscopic material property and recent approaches for the inversion of structure-property linkages identify whole sets of optimal structures. The proposed single-goal approach was therefore extended to form a *multi-equivalent-goal structure-guided processing path optimization* algorithm (MEG-SGPPO) for efficient optimization when multiple target-structures are given. We applied both algorithms to a simulated metal forming process and used the orientation density function (ODF) to represent the state of the microstructures (in this example, the crystallographic texture). To apply our generic algorithms to the exemplary application, we introduced a histogram-based ODF distance function.

In the result section of this paper, we report the results of a parameter study for the proposed ODF distance function and evaluation results of the algorithms SG-SGPPO and MEG-SGPPO applied to optimization problems on the metal forming process simulation. We show the ability of the multi-equivalent-goal algorithm to early prioritize promising microstructures from a set of multiple equivalent targets. We investigate the usefulness of the central features of both algorithms in the form of ablation study results.

The MEG-SGPPO algorithm generalizes over target-microstructures. This ability can further be used to transfer the learned process knowledge between applications. The transfer can be performed by using the trained Q-models across optimization tasks, which differ by varying sets of equivalent optimal microstructures. The application of deep reinforcement learning algorithms to sequences of tasks often suffers from *catastrophic forgetting* (knowledge about previous tasks is overwritten by recent training iterations). The incorporation of techniques like *elastic weight consolidation* (Kirkpatrick et al. 2017) into the multi-equivalent-goal algorithm could help to overcome this problem. SG-SGPPO and MEG-SGPPO depend on a core reinforcement learning algorithm. For the presented study we use the DQN algorithm with several enhancements. SG-SGPPO and MEG-SGPPO can be used with other Q-learning-based reinforcement learning methods than DQN, such as Neural Fitted Q-Iteration (Riedmiller 2005) or with actor-critic approaches, such as Deep Deterministic Policy Gradient (Lillicrap et al. 2015) to enable the optimization of processes with continuous control actions directly without discretization.

The proposed algorithms are not restricted to applications with orientation density functions as microstructural features or applications in metal forming only. Due to the model-free nature of the approaches, the algorithms can be applied to any new application field or process if (a) the material structure is observable during the process and (b) a distance function for the specific structure feature-space is defined.

The application of the proposed methods to the optimization of simulated processes with more accurate microstructure descriptors (2-point statistics, full-field models) furthermore requires an improved computational efficiency of the underlying process simulation. As the fields of data-driven computational material sciences and numerical simulation progresses rapidly, we are confident that applications that are intractable by now become feasible in the upcoming years.

Although we used a deterministic process simulation to evaluate the proposed methods, they are suitable also for non-deterministic processes. In particular, the multi-equivalent-goal prioritization is derived from the learned expectation functions. Thereby, our methods are also applicable for processes with stochastic state transitions. Notably, it is possible to learn directly on physical processes without the need for any prior process knowledge if the according structure features are measurable at processing time and a distance function on the structure features is defined.

As is typically the case with model-free reinforcement learning methods, the execution of the explorative learning policy follows no constraints and when used in combination with physical processes, the process safety has to be ensured by a proper constraint on the control quantities. The model-free algorithm builds its process knowledge during processing, which implies sub-optimality at the early stages of learning. This results in rejects in real processes or in excess computations in simulated processes. Such losses are significantly reduced by our multi-equivalent-goal approach, the data-augmentation, and the capability to generalize over spe-cific target-microstructures, which make the learning process much more data efficient.

**Acknowledgements** The authors would like to thank the German Research Foundation (DFG) for funding the presented work, which was carried out within the research project number 415804944: 'Taylored Material Properties via Microstructure Optimization: Machine Learning for Modelling and Inversion of Structure-Property-Relationships and the Application to Sheet Metals'. Also, we would like to thank Jan Pagenkopf for providing the crystal plasticity routine on which the implemented material model is based.

**Funding** Open Access funding enabled and organized by Projekt DEAL.